\documentclass[letterpaper, 10 pt, conference]{ieeeconf}  

\IEEEoverridecommandlockouts                              

\usepackage{graphics} 
\usepackage{epsfig} 
\usepackage{mathptmx} 
\usepackage{times} 
\usepackage{amsmath} 
\usepackage{amssymb}  
\usepackage{pifont}
\usepackage{hyperref}
\usepackage[dvipsnames]{xcolor}
\newcommand{\cmark}{\ding{51}}%
\newcommand{\xmark}{\ding{55}}%
\renewcommand{\baselinestretch}{0.974}

\title{\LARGE \bf
Deep Multi-view Depth Estimation with Predicted Uncertainty
}

\author{Tong Ke, Tien Do, Khiem Vuong, Kourosh Sartipi, and Stergios I. Roumeliotis$^{\dagger}$ 
\thanks{$^{\dagger}$Tong Ke, Tien Do, Khiem Vuong, Kourosh Sartipi, and Stergios I. Roumeliotis are with University of Minnesota, Minneapolis, MN 55455 {\tt\small \{kexxx069, doxxx104, vuong067, sarti009, stergios\}@umn.edu}.}
}

\begin{document}

\maketitle
\thispagestyle{empty}
\pagestyle{empty}

\begin{abstract}
In this paper, we address the problem of estimating dense depth from a sequence of images using deep neural networks.
Specifically, we employ a \textit{dense-optical-flow} network to compute correspondences and then triangulate the point cloud to obtain an initial depth map.
Parts of the point cloud, however, may be less accurate than others due to lack of common observations or small parallax.
To \textit{further} increase the triangulation accuracy, we introduce a depth-refinement network (DRN) that optimizes the initial depth map based on the image's contextual cues.
In particular, the DRN contains an iterative refinement module (IRM) that improves the depth accuracy over iterations by refining the deep features.
Lastly, the DRN also predicts the uncertainty in the refined depths, which is desirable in applications such as measurement selection for scene reconstruction.
We show experimentally that our algorithm outperforms state-of-the-art approaches in terms of depth accuracy, and verify that our predicted uncertainty is highly correlated to the actual depth error.

\end{abstract}

\section{Introduction}

Estimating dense depth from a sequence of images is necessary in applications such as 3D scene reconstruction and augmented reality.
Classical methods address this problem by first computing point correspondences based on hand-crafted matching criteria, and then constructing a 3D point cloud, given the camera pose estimates from structure-from-motion (SFM)~\cite{schonberger2016structure} or visual(-inertial) simultaneous localization and mapping (SLAM)~\cite{ORBSLAM, qin_tro_2018}.
They typically fail, however, to obtain reliable correspondences at low-texture or reflective surfaces, which leads to an incomplete scene reconstruction.

Recently, deep learning-based methods have shown the potential to compensate for the aforementioned limitation of the classical methods.
Specifically, approaches such as~\cite{Fu2018dorn,Li2018megadepth} predict dense depth from a single image by taking advantage of images' contextual cues learned from large datasets;
hence, they rely less on texture, as compared to classical methods.
Moreover, to overcome the scale issue of single-view methods, depth-completion networks (e.g., \cite{sartipi2020dde, teixeira2020aerial, wong2020voiced, qu2020depth, sinha2020deltas}) leverage sparse point clouds from classical methods and complete the dense depth map using single-view cues.
In order to further exploit multi-view information, depth-estimation networks taking multiple images as input have also been considered.
In particular, \cite{teed2020deepv2d, Liu2019neuralrgbd} employ cost volumes in their networks to embed geometric information, while \cite{xie2019video,luo2020consistent} explicitly leverage multi-view geometry by estimating dense optical flow.

In this work, we follow the latter paradigm.
Specifically, we employ an optical flow network to compute dense correspondences between a keyframe image and its immediate neighbors, and then triangulate the dense matches to compute the 3D point cloud given the cameras' poses.
Challenging conditions, however, such as lack of common observations or small parallax cause some points to have low-accuracy depth estimates.
To represent these errors, we employ the Hessian and residual of the triangulation least squares and define the confidence scores for the initially triangulated depths, which we then use in the depth-refinement network (DRN).

Although some of the aforementioned issues can be partially alleviated by applying an adaptive frame-selection policy (see Sec.~\ref{sec:exp}), we primarily focus on improving the accuracy by taking advantage of single-image depth estimation networks.
Specifically, in order to leverage the image's contextual information as well as the confidence scores, we introduce a DRN that uses as input the initially triangulated depths, their confidence scores, and the keyframe image to produce a refined depth map and its uncertainty (the aleatoric one~\cite{kendall2017uncertainties}).
In particular, we propose an iterative refinement module (IRM) in its decoder that iteratively refines the deep features extracted by the encoder using a least-squares layer, which significantly improves the depths' accuracy as compared to the initial ones.
The depth's uncertainty predicted by the DRN, however, and as shown by our experiments, are highly correlated with the actual depth errors, thus providing valuable criterion for selecting which measurements to use for 3D scene reconstruction. 
To summarize, our main contributions are:
\begin{itemize}
    \item We introduce an algorithm for estimating depth from multiple images, which outperforms state-of-the-art methods ($\sim$20\% lower RMSE on ScanNet~\cite{Dai2017scannet} dataset).
    \item We propose a depth-refinement network (DRN) with an iterative refinement module (IRM) that greatly improves the depths' accuracy and estimates their uncertainty.
    \item We further improve the depth estimation from multi-view by applying an adaptive frame-selection policy.
\end{itemize}

\begin{figure*}[t]
\includegraphics[width=\textwidth]{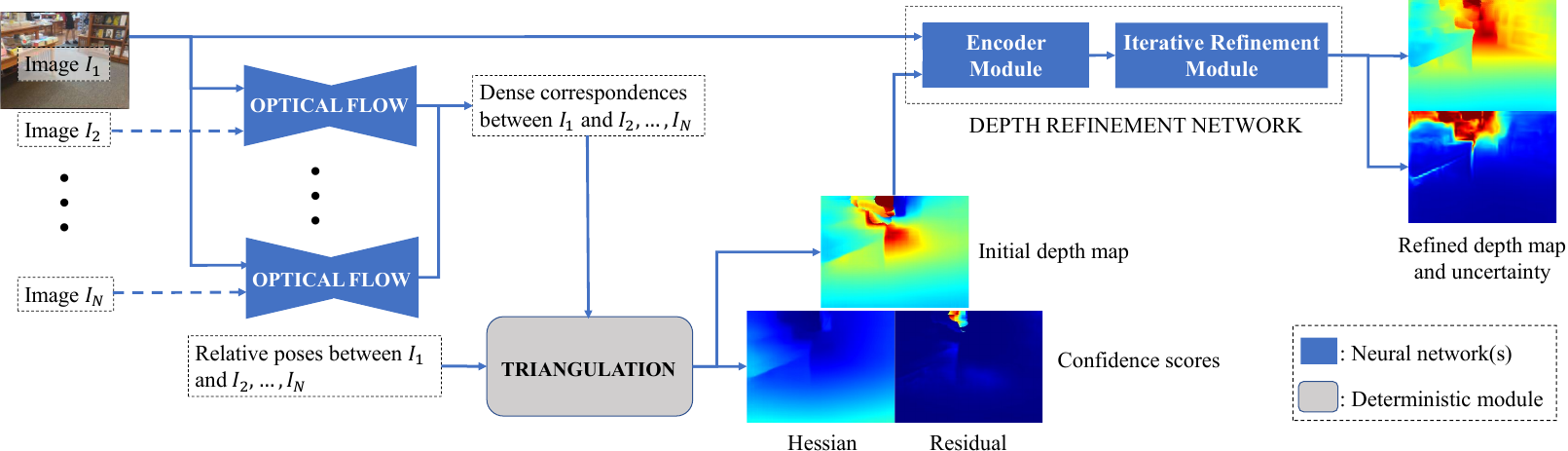} 
\centering
\caption{Overview of the system. For a keyframe $I_1$, Step 1: We compute the dense optical flow between image $I_1$ and $I_k, (k=2\dots N)$. Step 2: We triangulate the initial depth map of $I_1$ and compute its confidence scores (see Sec.~\ref{ssec:optical_flow}). Step 3: The DRN (Sec.~\ref{ssec:drn}) takes image $I_1$, the initial depth map, and confidence scores as input and iteratively [through its decoder (see Sect.~\ref{ssec:iterative})] outputs the refined depth map and 
its uncertainties. Note that the DRN includes an IRM (Sec.~\ref{ssec:iterative}) in its decoder, whose objective is to preserve initial depth estimates of high confidences scores.
}
\label{fig:overview}
\vspace{-5mm}
\end{figure*}

\section{Related Work}
Multi-view depth-estimation methods can be classified as:

\textbf{Depth completion:} One approach towards dense depth estimation from multiple views is to: (i)~Create a sparse point cloud (by tracking distinct 2D points across images and triangulating their 3D positions) and then (ii)~Employ a depth-completion neural network that takes the sparse depth image along with the RGB image as inputs and exploits the scene’s context to create a dense-depth estimate (e.g., \cite{sartipi2020dde, teixeira2020aerial, wong2020voiced, qu2020depth, sinha2020deltas}).
Although these approaches have relatively low processing requirements, they are typically sensitive to the inaccuracies and sparsity level of their depth input; thus, they often fail to produce accurate depth estimates in textureless regions that lack sparse depth information.
To overcome this limitation, we obtain a \textit{dense initial depth map} by triangulating a dense set of correspondences established by an optical flow method such as~\cite{teed2020raft}.
By doing so, and as we show experimentally, we significantly improve  accuracy as compared to sparse-to-dense depth completion approaches.
%

\textbf{Depth cost volume:} Alternatively, dense information from multiple frames is obtained by estimating a depth probability volume from depth cost volumes \cite{yao2018mvsnet, Liu2019neuralrgbd, Zhou2018deeptam, wei2019deepsfm, teed2020deepv2d, long2020occlusion}.
Specifically, by aggregating information across multiple frames, depth-cost-volume approaches yield higher accuracy compared to sparse-to-dense methods.
Their precision, however, is bounded by the range of their depth sweeping planes predefined in the cost volumes.
In contrast, our algorithm relies on optical flow and triangulation and thus, it is not restricted by the limited range and discretization effects of cost volumes.
In our experiments, we show that our method outperforms~\cite{teed2020deepv2d}, a state-of-the-art depth cost volume approach, by a significant margin ($\sim$20\% lower RMSE). 


\textbf{Flow-to-depth:} Lastly and closely related to our work is the approach of estimating dense depth from dense optical flow~\cite{Li2019ManneqinChallenge, xie2019video}.
Specifically, an initial depth map along with its confidence scores are first obtained through least-squares triangulation of the dense optical-flow correspondences.
Subsequently, the initial depth map is further improved by a DRN, often realized as a deep autoencoder, using its confidence scores and the RGB image. 
This DRN, however, may inaccurately modify the initial depth map, especially for pixels with high confidence scores (see Sect.~\ref{ssec:drn} for more details).
Previous works~\cite{Cheng2018cspn, xu2019depth} address this issue, in the context of depth completion when a \textit{sparse ground truth depth map} is given, by (i) replacing the refined depths with the ground truth ones when these are available, and (ii) propagating this information to the neighboring pixels via a convolution spatial propagation operator.
Our initial depth map, however, may contain significant noise and outliers, hence it cannot be employed as ground truth for depth propagation.

To address this issue, in this work, we improve the DRN by introducing an IRM that \textit{seeks to minimize the difference between the initial and final depth estimates of pixels with high confidence scores}.
To do so, we iteratively refine the joint deep feature representation of the RGB image, initial depth, and its confidence scores using a least-squares layer, analogous to the one in~\cite{hosseini2020dense, yaman2020self}.
%
%
As shown Sec.~\ref{sec:exp}, the proposed IRM leads to significant accuracy improvement as compared to simply using the confidence score as input to a neural network.
Furthermore, our approach estimates the refined depths’ uncertainty (aleatoric uncertainty model~\cite{ilg2018uncertainty, kendall2017uncertainties, poggi2020uncertainty}), which is employed to fuse dense-depth estimates across a scene~\cite{Liu2019neuralrgbd, teixeira2020aerial} (see 3D reconstruction experiment in Sect.~\ref{ssec:uncertainty}).

\section{Technical Approach}\label{sec:method}
We hereafter present our method for estimating the dense depth map of an image $I_1$ given $N-1$ adjacent images $I_2,\ I_3,\ \dots,\ I_N$ and their corresponding relative camera poses (these can be estimated online, e.g., by visual-inertial SLAM that combines inertial measurements with sparse visual feature tracks~\cite{KejianWuInverseFilter}).
Fig.~\ref{fig:overview} depicts an overview of our pipeline. Specifically, we first employ a dense optical flow network between images $I_1$ and $I_k, \ (k=2, \ \dots, \ N)$ to establish point correspondences and then use the cameras' relative poses for triangulation.
As a result, we obtain an \textit{initial depth map} for $I_1$, as well as the \textit{triangulation's confidence scores}.
Lastly, we employ a DRN that takes as input the initial depth map, the confidences scores, and image $I_1$ as input to refine the depth map and predict its uncertainty.
As mentioned earlier, in the DRN, we include an IRM which significantly increases the accuracy over iterations using a least-squares layer.
%
%
%
Next, we describe each module in detail.
\subsection{Optical Flow and Triangulation}\label{ssec:optical_flow}
For estimating optical flow, we employ the network of RAFT~\cite{teed2020raft} that takes as input a source image $I_1$ and a target image $I_k$ and computes a displacement $\delta\mathbf{u}_i$ for every pixel position ${}^{1}\mathbf{u}_i=\begin{bmatrix}  {}^{1}x_i & {}^{1}y_i & 1 \end{bmatrix}^T$ of $I_1$, so that ${}^{k}\mathbf{u}_i={}^{1}\mathbf{u}_i+\delta\mathbf{u}_i$ is its corresponding pixel in $I_k$.
Given a keyframe image $I_1$, for which we estimate the depth, and $N-1$ adjacent frames $I_2,\dots, I_N$, we run the optical-flow network pairwise between $I_1$ and $I_k$, $k=2,\dots, N$, so as to find the corresponding pixel in $I_k$ for every pixel of $I_1$.

From these correspondences and the relative poses of frames $I_1$ and $I_k$, $k=2,\dots, N$, we compute the initial depth $\bar{d}_i$ of each pixel in $I_1$ via triangulation. Specifically, we solve the following linear least-squares problem:
\begin{align}
    \bar{d}_i=\displaystyle\arg\min_{d_i}\displaystyle\sum_{k=2}^{N}\left\| (^{k}\mathbf{u}_i^\prime/\|^{k}\mathbf{u}_i^\prime\|)\times(^{k}_1\mathbf{R}{}^{1}\mathbf{u}_i^\prime d_i+{}^{k}\mathbf{p}_1)\right\|^2\label{eq:tri}
\end{align}
where $^{k}_1\mathbf{R}, \ {}^{k}\mathbf{p}_1$ are the orientation and position of frame $I_1$ with respect to frame $I_k$, and $^{n}\mathbf{u}_i^\prime\triangleq\mathbf{K}^{-1}{}^{n}\mathbf{u}_i,n=1\dots N$ with $\mathbf{K}$ being the camera intrinsic matrix.
Note that $^{k}\mathbf{u}_i$ may be outside of the image, while the triangulation still yields reasonable depths at those points, though typically less accurate.
As mentioned earlier, the initial depth map will be further improved by the DRN using as confidence scores the square root of the Hessian (which is a scalar here) and the norm of the residual from the least squares.
The former reflects the quality of triangulation, i.e., the parallax, while the latter represents the reprojection error.
Optionally, we select frames $I_2\dots I_N$ through an adaptive policy where $I_{k+1}$ is selected only if the relative rotation angle or distance between it and $I_{k}$ exceeds thresholds $t_r$, $t_d$, respectively. Using such policy, instead of using a time-based, fixed-size image sampling, improves the initial depth map's accuracy.

For training the optical flow network, a naive way is to apply an $\ell_1$ or $\ell_2$ loss on the depth computed from \eqref{eq:tri}.
This loss, however, only imposes constraints for the optical flow along the direction affecting the depth, i.e., the epipolar line, but not the direction perpendicular to it, where the magnitude of the triangulation residual is determined. To capture the errors in all directions, we propose the following loss by substituting the ground truth depth $d_i^*$ of pixel ${}^{1}\mathbf{u}_i$  to the least squares' cost function of a two-view triangulation:
\begin{align}
    l_{o}=\displaystyle\sum_{i}\left\|(^{2}\mathbf{u}_i/\|^{2}\mathbf{u}_i\|)\times(^{2}_1\mathbf{R}{}^{1}\mathbf{u}_i d_i^*+{}^{2}\mathbf{p}_1)\right\|^2\label{eq:lo}
\end{align}
%



\subsection{Depth Refinement Network (DRN)}\label{ssec:drn}
\begin{figure}[h]
\centering
\includegraphics[width=0.35\textwidth]{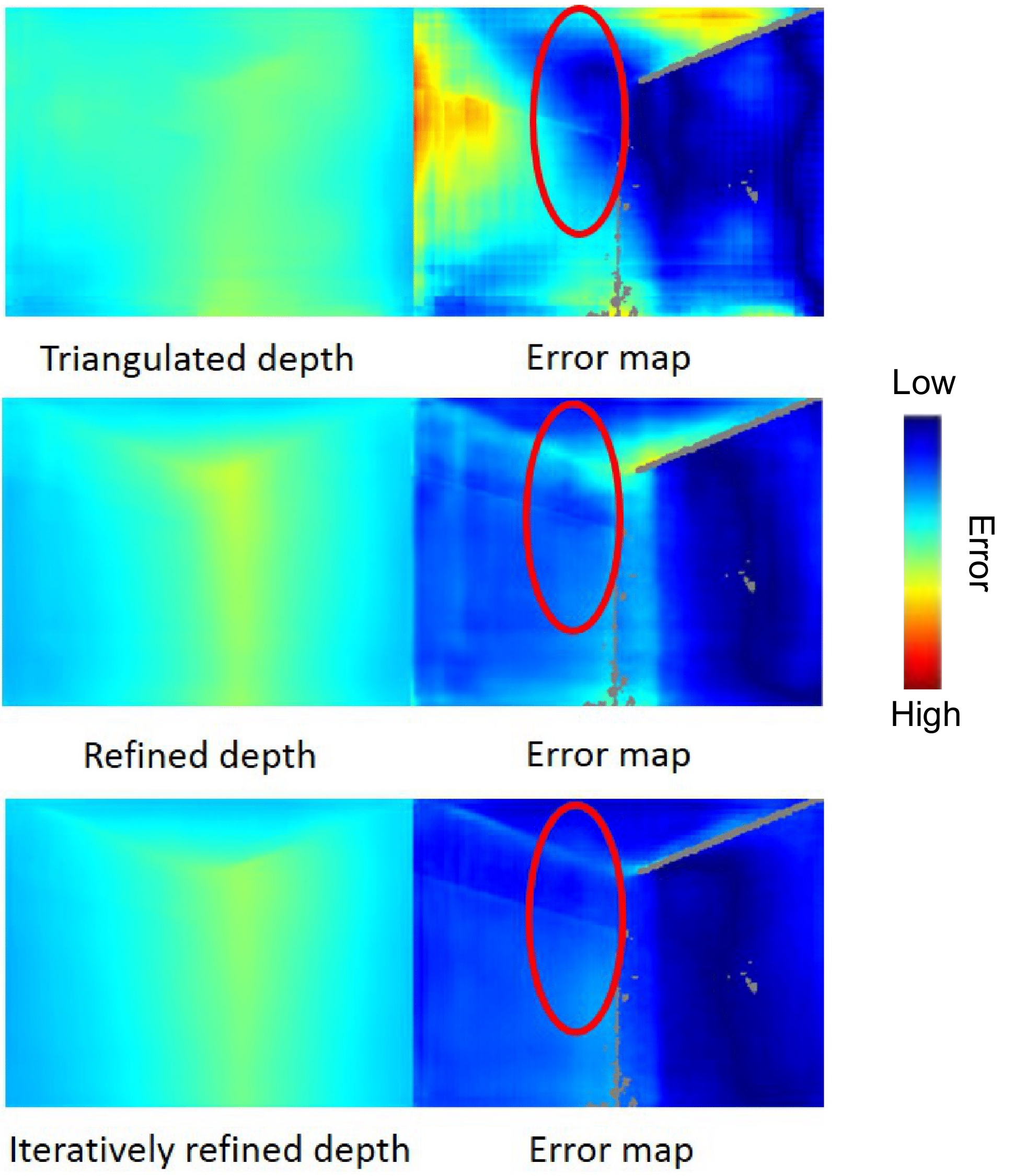}
\caption{Effect of the iterative depth refinement. The red ellipses indicate low-error regions in the triangulated depth. The initial step of the DRN causes the local error to increase but the iterative refinement module improves the depth estimates of these regions. The error color map uses red/blue for high/low error.}\label{fig:iterative_refinement}
\vspace{-5mm}
\end{figure}

The initial depth map and confidence scores (see Sect.~\ref{ssec:optical_flow}) are used by the DRN for further accuracy improvement.
%
%
In particular, the DRN seeks to preserve the initial triangulated depths for pixels with high confidence scores, while improving the rest of the depth map using the prior learned from the training data.
Previous works (e.g.,~\cite{Li2019ManneqinChallenge, xie2019video}) propose simply passing the initial depth map, its confidence scores, and the RGB image to an autoencoder to refine depth. 
As shown in Fig.~\ref{fig:iterative_refinement}, however, although the refined depths from an autoencoder network are overall more accurate than the triangulated depths, they are often incorrectly modified in the low-error regions.
To overcome this limitation, we employ the IRM described in Sect.~\ref{ssec:iterative} (see Fig.~\ref{fig:iterative_refinement}).
Lastly, note that the DRN approximates each output depth pixel as an independent Laplace random variable and the training of depth and uncertainty is performed with the aleatoric uncertainty model~\cite{kendall2017uncertainties, ilg2018uncertainty}.
The structure of DRN (Fig.~\ref{fig:overview}) comprises: (i)~An Encoder Module, and (ii)~An IRM. In what follows, we describe each module in details.

%
%

\begin{table*}[t]
\caption{Performance of depth prediction on ScanNet test set}
\vspace{-5mm}
\label{tab:scannet}
\begin{center}
\begin{tabular}{|c|ccccc|ccccc|}
\hline
 &    &      &     &     
 &          &                   &                   &   $E(\hat{d}, \delta)$       &                     &  \\ \cline{7-11}
 Method & Abs. Rel$\downarrow$ & Sq. Rel$\downarrow$ & log-RMSE$\downarrow$ & i-RMSE$\downarrow$ &
 RMSE $\downarrow$ & 1.05 $\uparrow$ &	1.10$\uparrow$  &	1.25$\uparrow$ &	$1.25^2 \uparrow$ &	$1.25^3 \uparrow$ \\
\hline
\hline
DDE-VISLAM~\cite{sartipi2020dde} & 0.156 &	0.079 &	0.174 &	0.134 &	
0.300 &	30.92 &	52.82 &	80.22 &	94.20 &	98.25\\
NeuralRGBD~\cite{Liu2019neuralrgbd} & 0.097	& 0.050	& 0.132	& 0.093	 & 
0.249 & - & - &	90.60 &	97.50 &	99.30\\
Flow2Depth~\cite{xie2019video} & 0.076 &	0.029 &	0.108	& 0.077	& 
0.199 & - & - &	93.30 &	98.40 &	99.60\\
DeepV2D~\cite{teed2020deepv2d} & 0.078  &	0.054  &	0.102  &	0.072 &
0.201  & 54.34 &	77.56 &	94.55 &	98.73 &	99.36\\
Ours (fixed) & \textbf{0.068} &	\textbf{0.026} &	\textbf{0.091} &	\textbf{0.064} &
\textbf{0.178} &	\textbf{57.38} &	\textbf{80.42} &	\textbf{95.75} &	\textbf{98.92} &	\textbf{99.61}\\
\hline
Ours (adaptive) & \textbf{0.058} &	\textbf{0.018} &	\textbf{0.082} &	\textbf{0.058} &	
\textbf{0.162} &	\textbf{62.63} &	\textbf{84.34} &	\textbf{96.77} &	\textbf{99.30} &	\textbf{99.77}\\
\hline
\end{tabular}
\end{center}
\vspace{-5mm}
\end{table*}

\subsubsection{Encoder Module}
%
%
This is employs an extension of the depth prediction architecture~\cite{sartipi2020dde}. Specifically, we first use three ResNet-18~\cite{He_2016_resnet} architectures to extract the deep-feature representations of the RGB image, the surface normal (predicted by a neural network from the RGB image as in~\cite{sartipi2020dde}), and the triangulated depth map with its confidence scores, and then concatenate them to generate a joint feature tensor $h$.
Given $h$, the estimated depth map $\hat{d}$ and its uncertainty $\hat{\sigma}$ are computed as:
\begin{align}
    \hat{d} = \mathbb{D} (h; \theta),  \
    \hat{\sigma} = \Sigma (h; \phi)
\end{align}
where $\theta$ and $\phi$ are the learned parameters of the depth decoder $\mathbb{D}$ and the uncertainty decoder $\Sigma$, respectively.
%
%
Next, we describe the IRM that further refines the feature $h$ to obtain a better depth estimate $\hat{d}$.

\subsubsection{Iterative Refinement Module (IRM)}\label{ssec:iterative}
The output of the depth decoder (a modified Panoptic Feature Pyramid Network~\cite{kirillov2019panoptic}) $\hat{d}$ from the initial step of DRN may contain erroneous estimates.
The IRM seeks to update the deep feature $h$ such that the difference between the estimated depth $\hat{d}$ and the initial depth estimate $\bar{d}$ becomes smaller for pixels with high confident values $\bar{c}$.
This can be formulated as a weighted least-squares problem, in which the weights $\hat{w}_i$ are computed from the deep feature $h$ via a weight decoder $\mathbb{W}$ with the learned parameters $\gamma$:
\begin{align}
\mathbb{C} (h)      &= \sum_{i} \hat{w}_i(\hat{d}_i-\bar{d}_i)^{2} = 
          \sum_{i} \mathbb{W}_i(h; \gamma)(\mathbb{D}_i(h; \theta)-\bar{d}_i)^{2} \label{eq:iter_cost_funct}
\end{align}
where $i$ indicates pixel position.
Note that $\hat{w}_i$ contains the joint information of the confidence scores and other inputs to the DRN.
%
We use a gated recurrent unit (GRU)~\cite{chung2014gru} to iteratively update $h$ as:
\begin{equation}
    h^{(k+1)} = \text{GRU}\left(h^{(k)}, \nabla_{h}\mathbb{C}\left(h^{(k)}\right)\right)
\end{equation}
where $\nabla_{h}\mathbb{C}(h)$ is computed using automatic differentiation~\cite{pytorch}. The reset and update gates inside each GRU learn to select the information between memory (deep feature $h^{(k)}$) and geometry (minimizing $\mathbb{C}(h)$ via $\nabla_{h}\mathbb{C}$).
Note that, for each iteration of the IRM, we only update the deep feature $h$, while we keep fixed the parameters of the GRU, the depth decoder $\mathbb{D}$, and the least-squares weights $\mathbb{W}$.
The whole network's parameters will later be updated through backpropagation with the ground-truth depths.
Lastly, the refined depth and its uncertainty are updated as:
\begin{align}
    \hat{d}^{(k+1)} = \mathbb{D}(h^{(k+1)}; \theta), \
    \hat{\sigma}^{(k+1)} = \Sigma (h^{(k+1)};\phi)
\end{align}

\begin{figure}[t]
\centering
\includegraphics[width=0.48\textwidth]{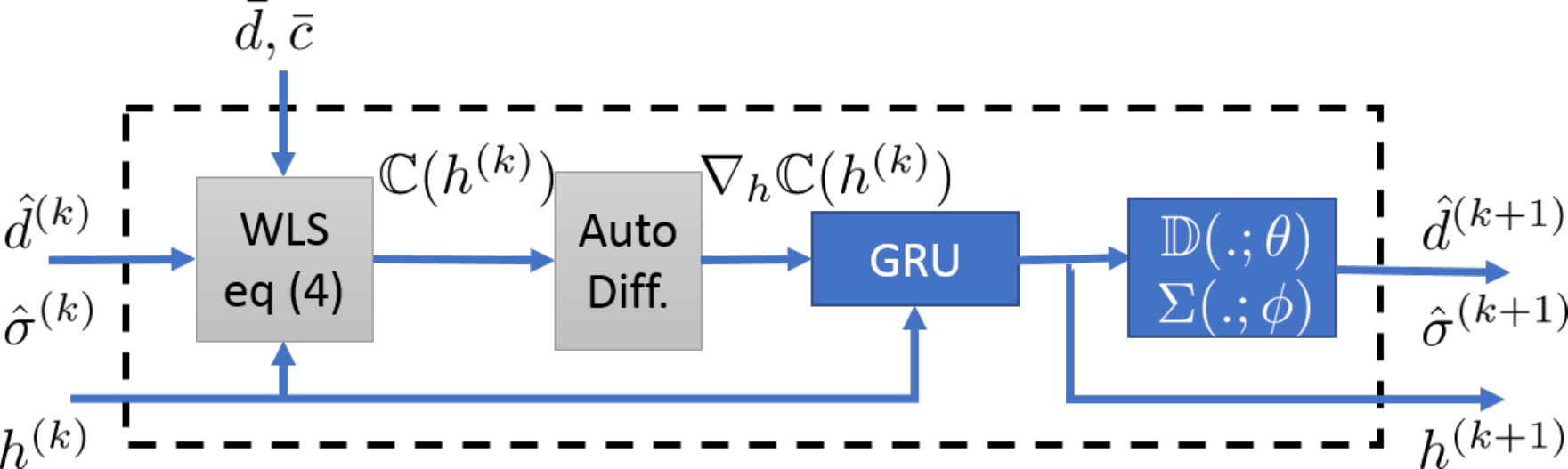}
\caption{An update block inside the IRM of Fig.~\ref{fig:overview}. At iteration $k$, it takes as input the feature $h^{(k)}$, estimated depth $\hat{d}^{(k)}$ at iteration $k$, and the optical-flow-based depth $\bar{d}$ along with its confidence $\bar{c}$, and outputs the updated feature $h^{(k+1)}$ and the updated estimated depth $\hat{d}^{(k+1)}$ and its confidence $\hat{\sigma}_{k+1}$.}\label{fig:update_block}
\vspace{-5mm}
\end{figure}

Fig.~\ref{fig:update_block} depicts the iterative refinement process for iteration~$k$.
Note that the learned parameters in the update block are shared across iterations.

During training, we execute the above optimization for a fixed number of $K$ iterations, where we employ the negative log-likelihood loss on the estimated depths with Laplace distribution~\cite{kendall2017uncertainties, ilg2018uncertainty}:

\begin{equation}
    l_r = \sum_{k=0}^{K} \lambda^{K-k}\left( \sum_{i} \frac{|\hat{d}_{i}^{(k)}-d_{i}^{*}|}{\hat{\sigma}_{i}^{(k)}} + \textrm{log }{\hat{\sigma}_{i}^{(k)}}\right) \label{eq:lr}
\end{equation}
with $d^{*}$ the ground truth depth and $\lambda<1$ a constant damping factor.
%
%
In our ScanNet experiment, we found through trial and error that $K=5$ and $\lambda=0.83$ offer balance between accuracy and efficiency.

\begin{figure*}[t]
\includegraphics[width=\textwidth]{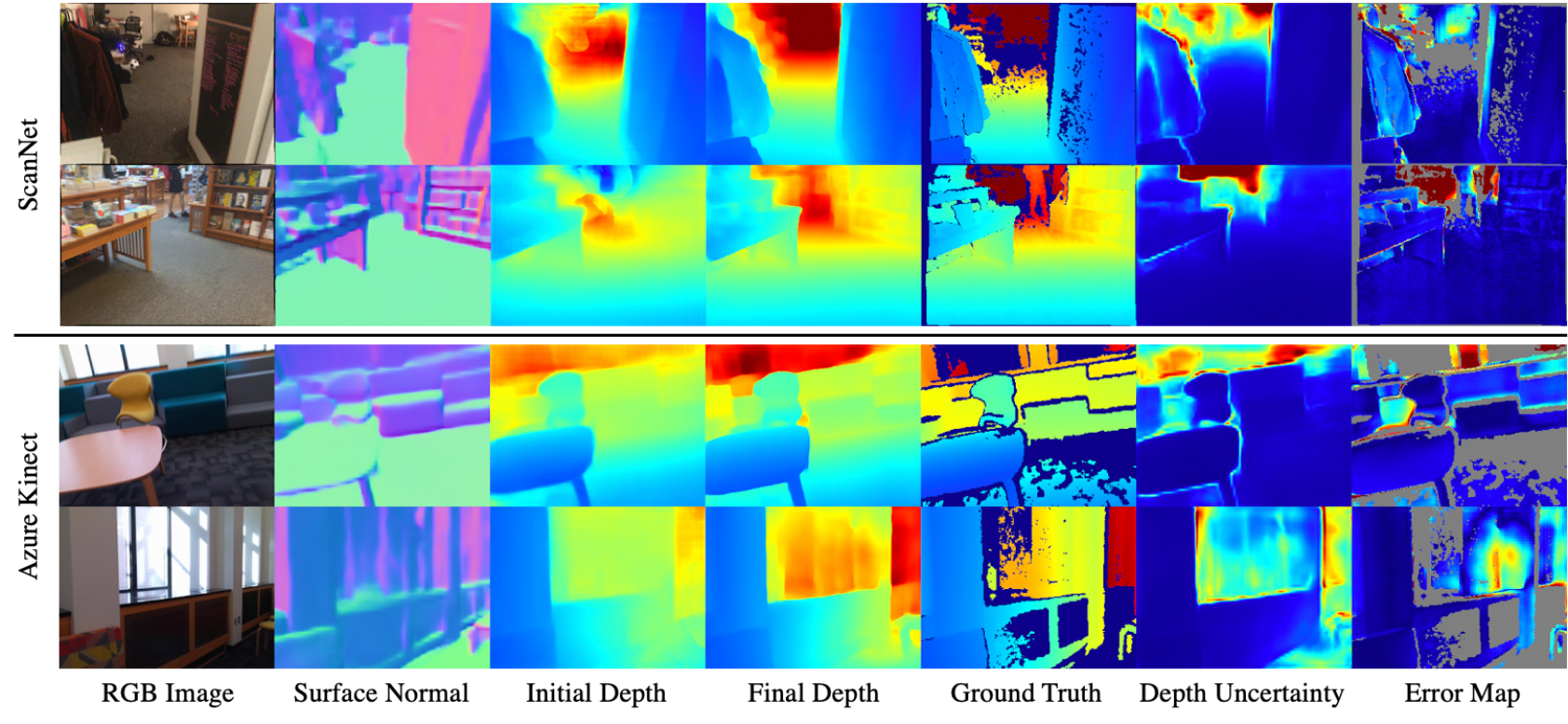} 
\centering
\vspace{-8mm}
\caption{Inputs, outputs, and intermediate results of our method. Note that the refined depths are significantly more accurate compared to the initial triangulated depths. Additionally, the uncertainty map is strongly correlated with the actual error map.}
\label{fig:qualitative_scannet}
\vspace{-3mm}
\end{figure*}
\section{Experimental Results}
\label{sec:exp}
In this section, we experimentally compare the performance of our method against state-of-the-art approaches and analyze the effect of the presented modules in ablation studies. 
To this end, we employ ScanNet~\cite{Dai2017scannet} and a dataset collected with Azure Kinect~\cite{KinectAzure} for evaluation.


\textbf{Evaluation metrics}: The accuracy of depths is assessed using multiple standard metrics, including: Mean absolute relative error (Abs. Rel); Mean square relative error (Sq. Rel); Logarithmic root mean square error (log-rmse); Root mean square error of the inverse depth values (iRMSE); Root mean square error (RMSE); and $E(\hat{d}, \delta)$ with $\delta=1.05,1.1,1.25,1.25^2,1.25^3$, defined as the percentage of the estimated depths $\hat{d}$ for which $\max(\frac{\hat{d}}{d^{*}},\frac{d^{*}}{\hat{d}}) < \delta$, where $d^{*}$ is the ground-truth depth.

\textbf{Experiment setup}: The networks in this paper have been implemented in PyTorch~\cite{pytorch} and our code is available at \url{https://github.com/MARSLab-UMN/DeepMultiviewDepth}.
The optical-flow network RAFT is trained using the loss in~\eqref{eq:lo} with the optimizer configuration of~\cite{teed2020raft}.
To train the DRN, we employ the Adam optimizer~\cite{kingma2015adam} with a learning rate of $10^{-4}$.
The training was done on an NVIDIA Tesla V100 GPU with 32GB of memory with a batch size of 16.
The results of other works are obtained by running the original authors' code with their provided network weights (when available).

\subsection{Comparison on ScanNet Dataset}\label{ssec:exp_scannet}
ScanNet~\cite{Dai2017scannet} is an RGB-D video dataset with more than 1500 sequences, annotated with 3D camera poses.
%
We employ the ScanNet standard training set to train our network and an evaluation set provided by~\cite{DORN} to assess our network's performance against the state-of-the-art competing approaches that are also trained on ScanNet:~\textit{DDE-VISLAM}~\cite{sartipi2020dde}\footnote{Note that the sparse input points in the evaluation set~\cite{DORN} are significantly fewer than the set used by \textit{DDE-VISLAM} for training; hence its performance is lower than the values reported in~\cite{sartipi2020dde}.}, \textit{NeuralRGBD}~\cite{Liu2019neuralrgbd}, \textit{Flow2Depth}~\cite{xie2019video}, and \textit{DeepV2D}~\cite{teed2020deepv2d}.
We employ the ground-truth poses for evaluation, and unless otherwise specified, five images with a fixed skipping interval of 5 are provided for depth estimation per each input sequence, where the depth is estimated for the middle (i.e.,~third) image.
%
For our algorithm, in addition to the fixed interval evaluation policy, denoted by \textit{Ours (fixed)}, we employ an adaptive frame-selection policy (see Sect.~\ref{ssec:optical_flow}) with $t_r=0.1 \ rad,t_d=0.08\ m$, denoted as \textit{Ours (adaptive)}.

Table~\ref{tab:scannet} summarizes the quantitative evaluation results of our proposed method and other state-of-the-art algorithms on the ScanNet.\footnote{Results of \textit{Flow2Depth} and \textit{NeuralRGBD} are obtained from~\cite{xie2019video}.}
As evident, \textit{Ours (fixed)} outperforms all alternative approaches with a clear margin in all evaluation metrics. 
We further improve the performance by employing the adaptive frame-selection policy [\textit{Ours (adaptive)}], featuring an overall $\sim$20\% decrease in RMSE as compared to \textit{DeepV2D} and \textit{Flow2Depth}.

Based on Fig.~\ref{fig:qualitative_scannet} (top two rows), which depicts qualitative results of our method on ScanNet,
we observe that: (i) The final refined depth map is closer to the ground truth as compared to the initial one, demonstrating the effectiveness of our depth refinement network;
(ii) The error map and the predicted uncertainty map (color-coded with the same scale) are highly correlated, which enables us to employ the latter for measurement selection and fusion (see  Sect.~\ref{ssec:uncertainty}).

Lastly, we measured the inference time to compute one depth map of \textit{DDE-VISLAM}, \textit{DeepV2D}, and \textit{Ours} on a single NVIDIA Tesla V100 GPU. \textit{DDE-VISLAM} takes, on average, only 44 ms, at the expense of lower accuracy. \textit{Ours} and \textit{DeepV2D} take 168 ms and 362 ms, respectively.

\subsection{Comparison on Azure Kinect dataset}\label{ssec:exp_azure}
To verify the generalization capability of our model, we perform cross-dataset evaluation by using the model trained on ScanNet to test on a dataset we collected with Azure Kinect~\cite{KinectAzure} containing $1528$ images.
We employ the depth-sensor data as the ground truth, while the sliding-window filter of~\cite{KejianWuInverseFilter} is used to estimate the camera poses. For \textit{Ours (fixed)} and DeepV2D, five images with a fixed skipping interval of 3 are provided for depth  estimation  per  each  input  sequence where  the  depth is estimated for the middle (i.e., third) image, while \textit{Ours (adaptive)} employs five images with adaptive policy.
%
Table~\ref{tab:azure} shows that \textit{Ours (fixed)} outperforms  \textit{DeepV2D}~\cite{teed2020deepv2d} in all metrics and \textit{DDE-VISLAM}~\cite{sartipi2020dde} in RMSE, 1.05, and 1.10. Moreover, the adaptive frame-selection policy employed by \textit{Ours (adaptive)} further improves the results and achieves the best accuracy in all metrics.
Fig.~\ref{fig:qualitative_scannet} (bottom two rows) depicts the qualitative results of our method on the Azure Kinect dataset.

\begin{table}[htb]
\caption{Performance of depth prediction on Azure Kinect test set.}
\vspace{-5mm}
\label{tab:azure}
\begin{center}
\resizebox{\columnwidth}{!}{
\begin{tabular}{|c|c|ccccc|}
\hline
 &        &           &   & $E(\hat{d}, \delta)$                       &  & \\ \cline{3-7}
Method & RMSE $\downarrow$ & 1.05 $\uparrow$ &	1.10$\uparrow$  &	1.25$\uparrow$ &	$1.25^{2}\uparrow$ &	$1.25^{3}\uparrow$\\
\hline
\hline
DDE-VISLAM~\cite{sartipi2020dde} &	0.298 &	28.75 &	50.46 &	\textbf{86.67} & \textbf{98.11} & \textbf{99.74} \\
DeepV2D~\cite{teed2020deepv2d} &	0.321 &	33.62 &	54.65 &	83.46  & 96.03 & 98.63\\
Ours (fixed) & \textbf{0.287} & \textbf{33.72} & \textbf{56.68} & 85.49 & 97.59	& 99.46\\
\hline
Ours (adaptive) & \textbf{0.265} & \textbf{35.97} &	\textbf{58.86} & \textbf{87.35} & \textbf{98.33} & \textbf{99.86} \\
\hline
\end{tabular}
}
\end{center}
\vspace{-5mm}
\end{table}


\begin{figure*}[t]
\includegraphics[width=0.95\textwidth]{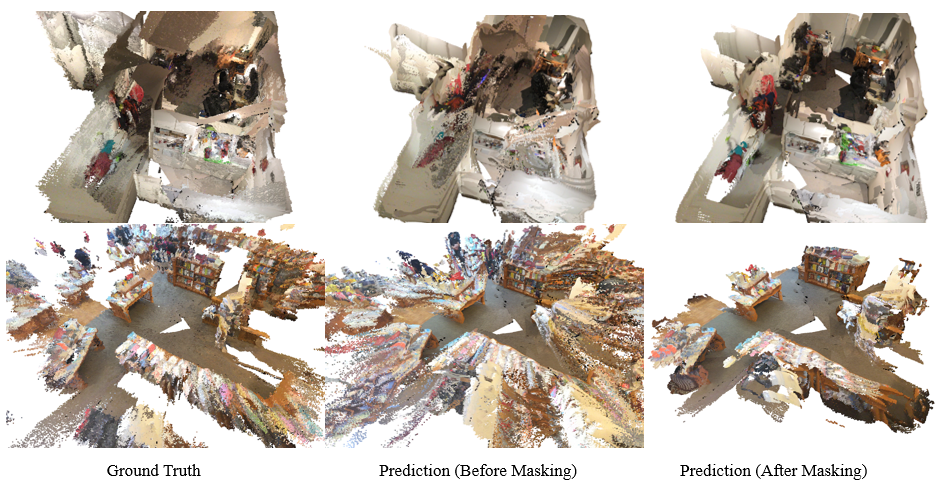} 
\centering
\vspace{-3mm}
\caption{3D scene reconstructions using the ground-truth, predicted, and masked depths.}
\label{fig:reconstruction_scannet}
\vspace{-3mm}
\end{figure*}

\subsection{Uncertainty Estimation}
\label{ssec:uncertainty}
\begin{table}[htb]
\caption{Depth accuracy with different uncertainty thresholds.}
\vspace{-8mm}
\label{tab:depth_uncertainty}
\begin{center}
\resizebox{\columnwidth}{!}{
\begin{tabular}{|c|c|c|ccc|}
\hline
&                   &                   &  & $E(\hat{d}, \delta)$      &    \\ \cline{4-6}
 $\hat{\sigma}$ thres. & Valid (\%) &	RMSE $\downarrow$ & 1.05 $\uparrow$ &	1.10$\uparrow$  &	1.25$\uparrow$ \\
\hline
\hline
 0.5 &	99.66 & 0.159 & 63.25 & 84.49 & 96.63 \\
 0.16 & 96.11 & 0.132 &	64.40 &	85.63 &	97.25 \\
 0.10 & 91.71 & 0.117 &	65.43 &	86.54 &	97.66 \\
 0.08 &	88.37 & 0.110 &	66.09 &	87.04 &	97.81 \\
\hline
\end{tabular}
}
\end{center}
\vspace{-5mm}
\end{table}

To verify the correlation between the predicted uncertainty estimates and the actual errors, we compare the depth error statistics when excluding points whose uncertainty estimates are larger than certain thresholds as shown in Table~\ref{tab:depth_uncertainty} [using results of \textit{Ours (adaptive)}].
%
Decreasing the value of the acceptable predicted uncertainty $\hat{\sigma}$ results in more accurate depth estimates at a small loss of image coverage.
For example, we retain 90\% of the depth values and reduce the RMSE by $\sim$20\% when excluding points with uncertainty above $0.1$.
Additionally, the impact of the uncertainty-based masking of the predicted depth images on scene reconstruction (from the ScanNet dataset) is depicted in Fig.~\ref{fig:reconstruction_scannet}, where the depth RMSE is reduced by more than 1.4 (3.1) times for the scenes in the top (bottom) row, while only removing 20\% of the total depth estimates.
Hence, we demonstrated quantitatively and qualitatively that the uncertainty-based depth masking improves reconstruction accuracy.

\subsection{Ablation Study}\label{ssec:ablation}

In this section, we analyze each component of our pipeline that contributes to the overall performance gain (19\% in RMSE as compared to other state-of-the-art methods). 

\textbf{Sparse vs. Dense}: In the proposed method, a \textit{dense} initial depth map is provided to the DRN, instead of a sparse one as in depth-completion approaches.
In order to study the effect of the initial depth map's density, we compare our DRN, without the IRM in its decoder [\textit{Ours~(w/o IRM)} ; see Sect.~\ref{ssec:drn}], to the DDE-VISLAM~\cite{sartipi2020dde}, a depth completion network that takes sparse depth as input.
Specifically, we randomly sample a fixed number of sparse points (e.g., 10, 100, 200 in Table~\ref{tab:sparse_dense}) from the initial triangulated depth map that have high confidence scores (see Sect.~\ref{ssec:optical_flow}).
These sampled depths are added to the sparse depth input of the \textit{DDE-VISLAM}.
In Table~\ref{tab:sparse_dense}, we show that providing the dense depth estimates together with confidence scores contributes $\sim$14.5\% in RMSE improvement as compared to the depth completion (\textit{DDE-VISLAM+200}) and other state-of-the-art approaches.\footnote{Note that \textit{DDE-VISLAM+200} performs comparable, in terms of RMSE, to the state-of-the-art methods \textit{DeepV2D} and \textit{Flow2Depth}.}
These results confirm our hypothesis that employing a dense, instead of a sparse, initial depth map improves accuracy.

\begin{table}[htb]
\vspace{-3mm}
\caption{Depth accuracy with sparse and dense input}
\vspace{-8mm}
\label{tab:sparse_dense}
\begin{center}
\resizebox{\columnwidth}{!}{
\begin{tabular}{|c|c|c|ccc|}
\hline
&                   &                   &  & $E(\hat{d}, \delta)$      &    \\ \cline{4-6}
 Method & Dense &	RMSE $\downarrow$ & 1.05 $\uparrow$ &	1.10$\uparrow$  &	1.25$\uparrow$ \\
\hline
\hline
 DDE-VISLAM &	\xmark & 0.300 &	30.92 &	52.82 &	80.22 \\
 DDE-VISLAM + 10 & \xmark & 0.218 &	45.92 &	70.92 &	92.21 \\
 DDE-VISLAM + 100 &	\xmark & 0.201 & 51.23 & 75.72	& 93.76 \\
 DDE-VISLAM + 200 &	\xmark & 0.200 & 51.43 & 75.91	& 93.79 \\
 \hline
 Ours (w/o IRM) &	\cmark & \textbf{0.171} &	\textbf{59.25} &	\textbf{82.32}	& \textbf{96.16} \\
\hline
\end{tabular}
}
\end{center}
\vspace{-3mm}
\end{table}

\textbf{Iterative vs. Non-iterative}: To demonstrate the effectiveness of the proposed IRM (Sect.~\ref{ssec:iterative}), we compare our DRN with and without it, denoted as \textit{Ours~(w/~IRM)}, and \textit{Ours~(w/o~IRM)}, respectively.
During training, we use five iterations, while at inference time, we use seven iterations.
Table~\ref{tab:iter_vs_noniter} shows that the IRM contributes an additional $4.5\%$ improvement in RMSE.
Furthermore, we observe that there is little improvement after seven iterations, hence we limit the number of refinement steps during inference accordingly.

\begin{table}[htb]
\caption{Depth accuracy with IRM}
\vspace{-8mm}
\label{tab:iter_vs_noniter}
\begin{center}
\resizebox{\columnwidth}{!}{
\begin{tabular}{|c|c|c|ccc|}
\hline
&                   &                   &  & $E(\hat{d}, \delta)$      &    \\ \cline{4-6}
 Method & Iterations &	RMSE $\downarrow$ & 1.05 $\uparrow$ &	1.10$\uparrow$  &	1.25$\uparrow$ \\
\hline
\hline
 Ours (w/o IRM) &	0 & 0.171 &	59.25 &	82.32	& 96.16 \\
 \hline
  & 1 & 0.166	& 61.60	& 83.66	& 96.55 \\
  & 3 & 0.163	& 62.47	& 84.20	& 96.72 \\
  Ours (w/ IRM) & 5 & \textbf{0.162} &	62.59 &	84.30 &	96.76 \\
 & 7 & \textbf{0.162}	& 62.63	& 84.34	& 96.77 \\
  & 9 & \textbf{0.162}	& \textbf{62.64}	& \textbf{84.35}	& \textbf{96.78} \\
\hline
\end{tabular}
}
\end{center}
\vspace{-3mm}
\end{table}

\begin{table}[htb]
\caption{Depth accuracy with different input confidence scores}
\vspace{-8mm}
\label{tab:input_confidence}
\begin{center}
\resizebox{\columnwidth}{!}{
\begin{tabular}{|c|c|c|ccc|}
\hline
&                   &                   &  & $E(\hat{d}, \delta)$      &    \\ \cline{4-6}
 Configuration & Param.s (M) &	RMSE $\downarrow$ & 1.05 $\uparrow$ &	1.10$\uparrow$  &	1.25$\uparrow$ \\
\hline
\hline
 $\bar{d}$ &	39.241 &	0.168 &	61.79 &	83.60 &	96.28 \\
 $\bar{d},\bar{c}_r$ & 39.244 &	0.164 &	62.61 &	84.33 &	96.48 \\
 $\bar{d},\bar{c}_h$ & 39.244 &	0.168 &	61.98 &	83.39 &	96.35 \\
 $\bar{d},\bar{c}$ & 39.247 & \textbf{0.162} &	\textbf{62.63} &	\textbf{84.34} &	\textbf{96.77} \\
\hline
\end{tabular}
}
\end{center}
\vspace{-8mm}
\end{table}
\textbf{Importance of the triangulation confidence scores:} To assess the significance of the initial depths' confidence scores (Sec.~\ref{ssec:optical_flow}), we train our DRN with the IRM as previously described, with four different input options: (i)~The triangulated depth map ($\bar{d}$); (ii) The triangulated depth map and the residual as confidence score ($\bar{d},\bar{c}_r$); (iii) The triangulated depth map and the square root of the Hessian as confidence score ($\bar{d},\bar{c}_h$); (iv) The triangulated depth map and both confidence scores ($\bar{d},\bar{c}$), $\bar{c}=\{\bar{c}_r, \bar{c}_h \}$. As evident from Table~\ref{tab:input_confidence}, (iv) yields the best result with less than $0.02\%$ increase in parameters [see Param.s column in (M) millions].

\section{Conclusions and Future Work}


In this paper, we introduced an algorithm that employs multiple image to compute the dense depth representation of scene with the corresponding uncertainty.
Specifically, pixels tracked by dense optical flow are triangulated and provided to a depth refinement network (DRN) that further improves depth-estimation accuracy.
To do so, the DRN first extracts deep features from the inputs and then performs a neural least-squares optimization within its iterative refinement module.
In addition to the depth estimates, their corresponding uncertainty is predicted which is shown experimentally to be highly correlated with the actual depth errors.
As part of our future work, we plan to employ the estimated uncertainty for selecting the depth measurement to be used in global 3D scene reconstruction.

\clearpage



\bibliographystyle{IEEEtran}
\bibliography{references}

\end{document}